\documentclass{article}
\usepackage{graphicx} 

\title{A Directional Rockafellar-Uryasev Regression}
\author{Alberto Arletti}
\date{May 2024}

\usepackage{latexsym}
\usepackage{amssymb}
\usepackage{amsmath}
\usepackage{amsthm}
\usepackage{booktabs}
\usepackage{enumitem}
\usepackage{graphicx}
\usepackage{color}
\usepackage{tikz}

\usepackage[numbers]{natbib}
\usepackage{authblk} 

\author[1]{Alberto Arletti}
\affil[1]{Department of Statistical Science, University of Padua, Italy.}

\usetikzlibrary{shapes.geometric}

\DeclareMathOperator*{\argmin}{arg\,min}

\begin{document}

\maketitle

\section*{Abstract}
Most ost Big Data datasets suffer from selection bias. For example, X (Twitter) training observations differ largely from the testing offline observations as individuals on Twitter are generally more educated, democratic or left-leaning. Therefore, one major obstacle to reliable estimation is the differences between training and testing data. How can researchers make use of such data even in the presence of non-ignorable selection mechanisms? A number of methods have been developed for this issue, such as distributionally robust optimization (DRO) or learning fairness. A possible avenue to reducing the effect of bias is meta-information. Researchers, being field exerts, might have prior information on the form and extent of selection bias affecting their dataset, and in which direction the selection might cause the estimate to change, e.g. over or under estimation. At the same time, there is no direct way to leverage these types of information in learning. I propose a loss function which takes into account two types of meta data information given by the researcher: quantity and direction (under or over sampling) of bias in the training set. Estimation with the proposed loss function is then implemented through a neural network, the directional Rockafellar-Uryasev (dRU) regression model. I test the dRU model on a biased training dataset, a Big Data online drawn electoral poll. I apply the proposed model using meta data information coherent with the political and sampling information obtained from previous studies. The results show that including meta information improves the electoral results predictions compared to a model that does not include them.

\section{Introduction}

A major challenge has to be faced when training any model: the possibility of training data $P$ being non-representative of the overall population $Q$. In this sense, the estimated model $P_P(Y|X)$ might be different from the true model $P_Q(Y|X)$ due to some bias. Bias might be induced by a selection mechanism, $S \in \{0, 1\}$, such that $P_Q(Y|X,S=1) = P_P(Y|X)$. For example, Big Data can notoriously suffer from selection mechanisms which might induce bias. Notoriously, data drawn form the web are not representative of the overall population \citep{meng2018statistical}, \citep{baker2013summary} \citep{zagheni2015demographic}. In this sense, due to a survey being conducted online, the selection mechanism can affect the estimation. As an example, participants might have different characteristics from the general population, such as higher income, or a preference for the politically left among others \citep{giorgi2022correcting}. These unmeasured characteristics in turn changes the relationship between the target variable and the covariates, acting as a confound \citep{disogra2011calibrating}. One other example of this issue is the case of electoral polls. Many electoral polls recruit large samples, similarly to Big Data, but this sample might be drawn from online panel of respondents \citep{mcpheedata} or through social media questionnaires \citep{wang2015forecasting}. In both cases, the information drawn form the sample might present a scenario different substantially from the overall population \citep{callegaro2014critical}. One crucial challenge in machine learning is therefore to draw useful inferences when the data at hand contain some selection bias. It should be noted that this problem is not only limited of the social sciences, for biased training sets are often encountered in other fields of machine learning \citep{zhou2022domain, liu2024need, cai2023diagnosing, shen2020stable, kim2019learning}. In the machine learning literature bias-induced changes in $P(Y|X)$ are also referred to as distribution shifts, such as $X$-shifts and $Y|X$-shifts \citep{koh2021wilds}. Again, this problem it closely interconnected with the topic of Distributionally Robust Optimization (DRO), where a model is trained to reduce loss among a range of possible target distributions (robustness set) to account for shifts. Nonetheless, the aforementioned problem might also be phrased in therm of a missing data problem. In that case, to each observation in the sample is assigned $S = 1$ and the rest of the target population is considered unobserved with $S = 0$. One important theoretical framework of missing data problems is types of missingness, divided into Missing Completely at Random (MCAR) or Missing at Random (MAR), and Missing Non at Random (MNAR) \citep{rubin1976inference}. In the first case we have that $P(S=1|X,Y) = P(S|X)$ or $P(Y|S=1,X) = P(Y|X)$, while in the second case $P(S|X,Y)$ or $P(Y|S,X)$ cannot be reduced to $P(S|X)$ or $P(Y|X)$. In other words, MNAR samples contain some selection bias which makes inference unreliable. Continuing, the same concept can also be phrased in therms of the Data Defect Index (DDI) \citep{meng2018statistical}. DDI indicates the correlation between the sampling mechanism $S$ and the target variable $Y$. The expected DDI for MNAR (non-probability or non-random) samples is different from 0 drastically reducing the effective sample size and \citep{meng2022comments}. The appearance of this topic across the social sciences, machine learning or survey science is a testament to its relevance.

\section{Previous work}

\begin{figure}[h]
    \centering
    \includegraphics[width=\linewidth]{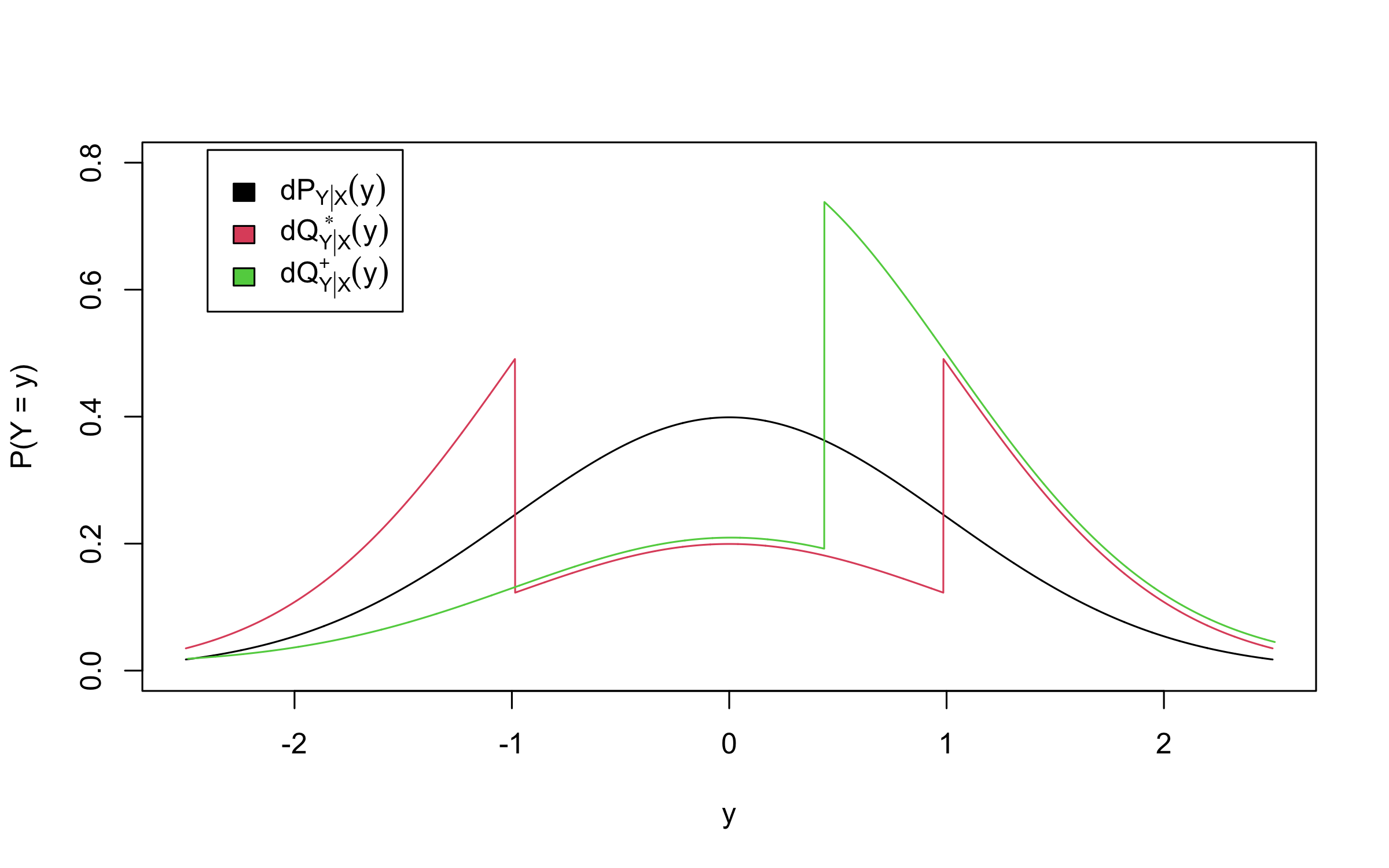}
    \caption{Representation of worst case distribution in RU and dRU robustness sets}
    \label{fig:worst_case}
\end{figure}

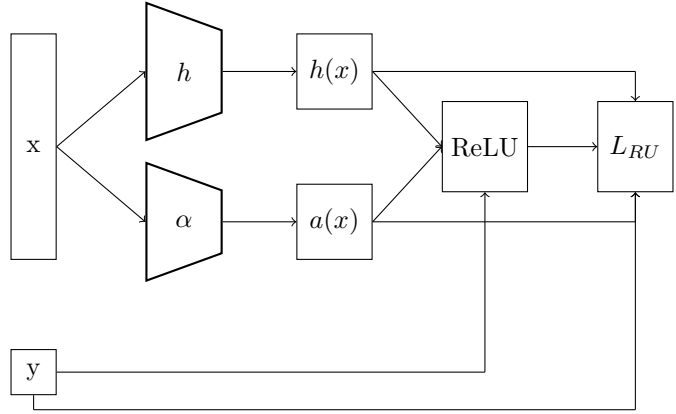
\begin{figure}
    \centering

\begin{tikzpicture}

\node (x) at (0,0) [draw, minimum height=3cm, minimum width=0.6cm] {x};
\node (y) at (0,-3) [draw, minimum height=0.6cm, minimum width=0.6cm] {y};
\node (h) at (2,1) [trapezium, trapezium angle=70, minimum width=0.6cm, minimum height=1cm, draw, thick, rotate=-90] {\rotatebox{90}{$h$}};;
\node (a) at (2,-1) [trapezium, trapezium angle=70, minimum width=0.6cm, minimum height=1cm, draw, thick, rotate=-90]{\rotatebox{90}{$\alpha$}};
\node (hx) at (4,1) [draw, minimum height=1cm, minimum width=1cm] {$h(x)$};
\node (ax) at (4,-1) [draw, minimum height=1cm, minimum width=1cm] {$a(x)$};
\node (ReLU) at (6,0) [draw, minimum height=1.2cm, minimum width=1cm] {ReLU};
\node (LRU) at (8,0) [draw, minimum height=1.2cm, minimum width=1cm] {$L_{RU}$};
\node (anch0) at (0,-3.5) {};
\node (anch1) at (6,-3) {};
\node (anch2) at (8,1) {};
\node (anch3) at (8,-1) {};
\node (anch4) at (8,-3.5) {};

\draw[->] (x.east) -- (h.south);
\draw[->] (x.east) -- (a.south);
\draw[-] (y.east) -- (anch1.center);
\draw[-] (y.south) -- (anch0.center);
\draw[-] (anch0.center) -- (anch4.center);
\draw[->] (anch4.center) -- (LRU.south);
\draw[->] (anch1.center) -- (ReLU.south);
\draw[->] (h.north) -- (hx.west);
\draw[->] (a.north) -- (ax.west);
\draw[->] (hx.east) -- (ReLU.west);
\draw[->] (ax.east) -- (ReLU.west);
\draw[->] (ReLU.east) -- (LRU.west);
\draw[-] (hx.east) -- (anch2.center);
\draw[->] (anch2.center) -- (LRU.north);
\draw[-] (ax.east) -- (anch3.center);
\draw[->] (anch3.center) -- (LRU.south);

\end{tikzpicture}
    \caption{Schematic representation of an implementation of RU or dRU regression models with neural networks.}
    \label{fig:ru-schema}
\end{figure}

Possible approaches to adjust estimation for discrepancies between sample and target population exist in the literature as long as the $S$ mechanism is MAR. For example, weighting can effectively be used to re-weight the sample to the general population using the $X$ available covariates, such as in inverse probability weighting (IPW) \citep{chen2020doubly}. The important assumption of IPW, and similar methods, is that the inclusion probability weights $\pi_i$ of each observations can be estimated correctly from the covariates $X$. There is no guarantee for this assumption to be respected in the case of MNAR. The same can be said for similar methods such as post-stratification \citep{si2020use}. To relax this assumption, \citep{aronow2013interval} assumes the weight cannot be precisely estimated from the sample but can be expected to be included in boundaries $ \alpha \leq \pi_i \leq \beta $. In other words, the sample inclusion weights suffer from a certain amount of bias bounded by $\gamma = \alpha / \beta$. Then, estimates of the target variable mean can be drawn considering the amount of bias. This approach has been furthered by \citep{sahoo2022learning}, which consider the case where a dataset suffer from bias quantified by $\Gamma$, where $\Gamma$ amounts degree of separation of $S$ from MAR: \begin{align*}
    \frac{P(S=1|X, Y)}{P(S=1|X)} \in \{\Gamma^{-1}, \Gamma\}.
\end{align*}
In this sense regression $\Gamma$ represents meta-information about the total quantity of bias to which the data is subjected to. The authors define Gamma-biased sampling a sampling process from a population $Q$ such as the sample $P$ results with a bias of $\Gamma$, such as follows: Let $P$, $Q$ be the distributions over $(X, Y)$. $Q$ can generate $P$ via $\Gamma$-biased sampling if and only if \begin{align*}
    \Gamma^{-1} \leq \frac{dQ_{Y|X}(y)}{dP_{Y|X}(y)} < \Gamma
\end{align*}. 
In this sense, $\Gamma$ represents the most extreme value of the density ratio between the sampled and target distributions. The authors present an approach to obtain an estimating function $h(x)$ so that it minimizes a loss function $L(h(x), y)$ of the worst case among a robustness set of distributions defined as $\mathcal{S}_{\Gamma} (P, Q_X)$, where $Q$, the population, can generate $P$, the sample, under Gamma-biased sampling. The minimization task is presented as: 
\begin{align} \label{eq:task}
    \min_{h(x)} \sup \big\{ \mathbb{E}_{Q_{Y | X}} [ L(h(x), Y) | X = x ] : Q \in \mathcal{S}_{\Gamma}(P, Q_X) \big\}.
\end{align}
A convex function over a closed, bounded, convex set is maximized at an extremal point of the set \citep{rockafellar1997convex}. Therefore, the loss is maximized by its worst-case distribution among the robustness set. The worst case distribution will be the distribution which returns the highest loss given $L$, when $h(x)$ is trained on $P$. An example of a worst-case distribution for the squared-loss when $P$ is a normal distribution is presented as $dQ^*_{Y|X}(y)$ in Figure \ref{fig:worst_case}, and consists in a symmetric heavy tailed distribution with the same mean. Through the use of the properties of Conditional Value at Risk \citep{rockafellar2000optimization}, the authors propose an approach to solve \ref{eq:task}, named Rockafellar-Uryasev (RU) regression. RU regression can be estimated via a neural network (see Figure \ref{fig:ru-schema} for a schematic representation) and the following loss function: 
\begin{align*}
    & (h^*_{\Gamma}, \alpha^*_{\Gamma}) = \argmin_{h, \alpha} \mathbb{E}_P [L_{RU}^{\Gamma}(h(X), \alpha(X), Y) ] \\ 
    & L_{RU}^{\Gamma}(z, a, y) = \Gamma^{-1} L(z, y) + (1 - \Gamma^{-1}) a \\ 
    & + (\Gamma - \Gamma^{-1})(L(z, y) - a)_{+}.
\end{align*}

\begin{figure}[h]
    \centering
    \includegraphics[trim=0 0 0 80, clip, width=\linewidth]{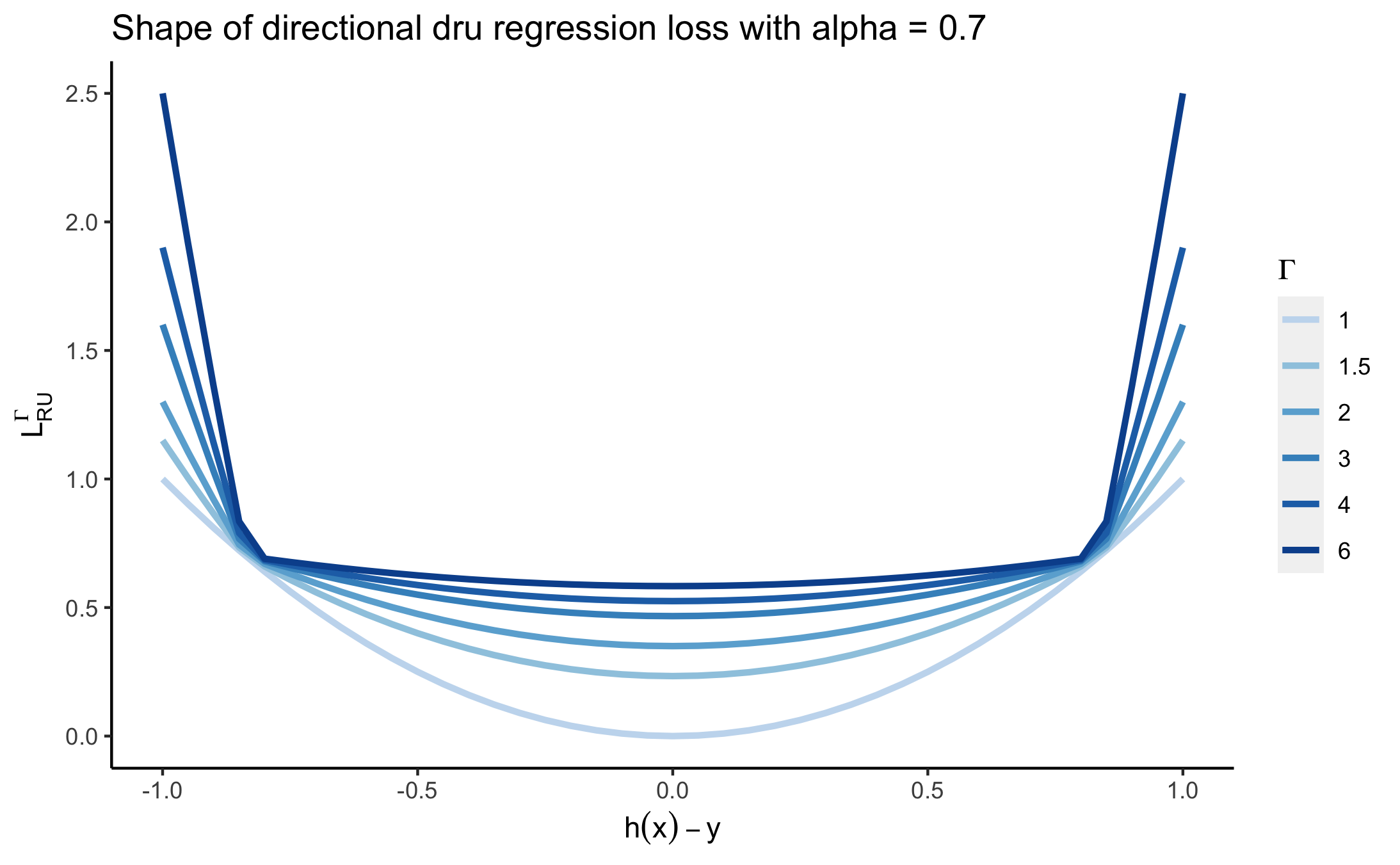}
    \caption{Visualization of RU regression loss for different values of $\Gamma$}
    \label{fig:loss_viz_gamma}
\end{figure}

\begin{figure}[h]
    \centering
    \includegraphics[trim=0 0 0 80, clip, width=\linewidth]{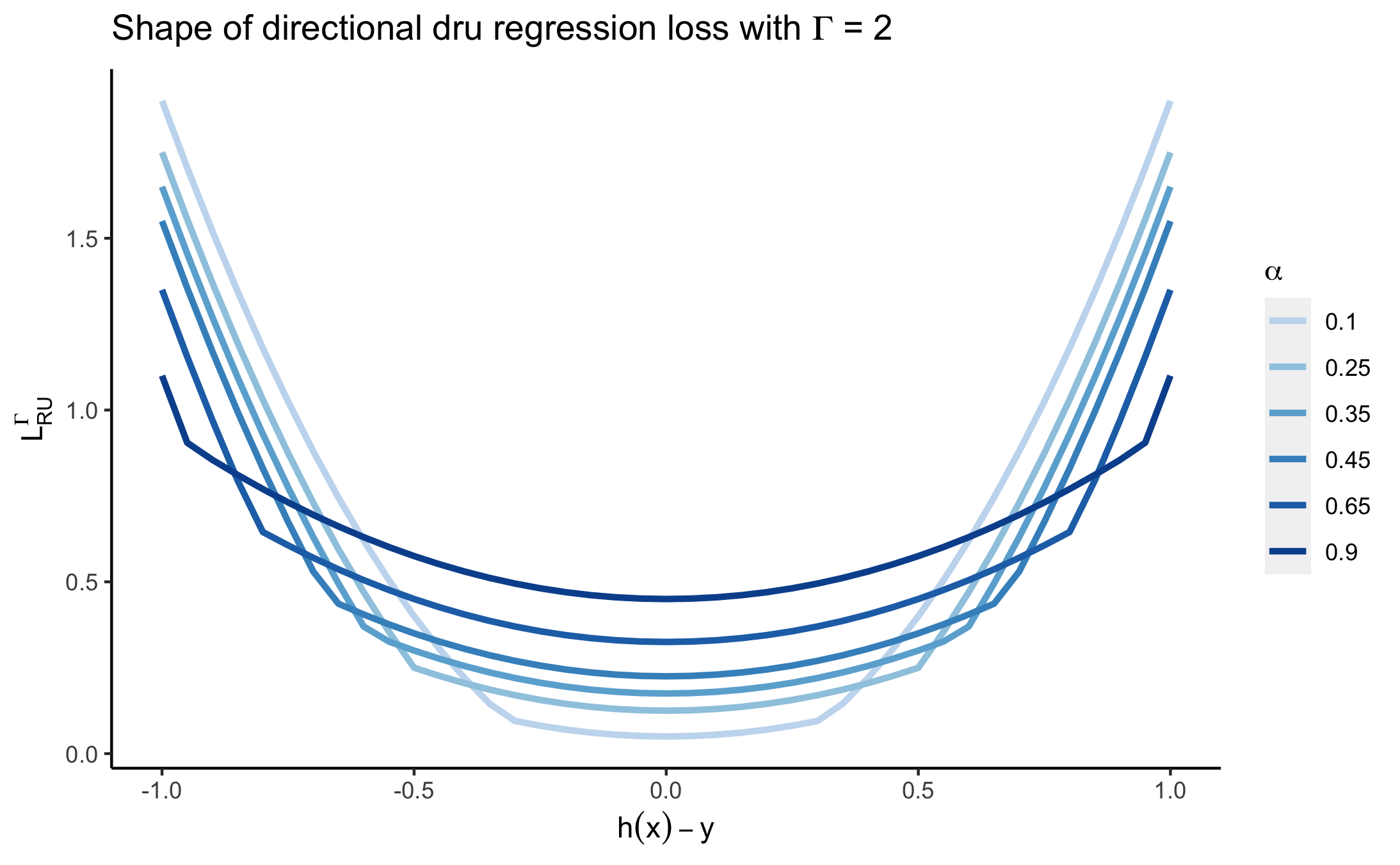}
    \caption{Visualization of RU regression loss for different values of $a$}
    \label{fig:loss_viz_alpha}
\end{figure}

The loss function takes as input the output of two neural networks, $h(x)$ the predictor network, and $\alpha(x)$ an auxiliary network used to distribute the penalization. For a schematic visualization of the loss function, see Figures \ref{fig:loss_viz_gamma} and \ref{fig:loss_viz_alpha}. The value of $\Gamma$ makes larger losses more costly, so that the network will tend to avoid any large loss during training, for higher values of $\Gamma$. 

One consequence of the shape of the loss function and of the way the worst case distribution is chosen in RU regression is that as $\Gamma$ increases the estimates will tend to approach $(\max(y) - \min(y))/2$. This is due to the fact that as $\Gamma$ increases, larger losses are more penalized compared to smaller losses. Therefore, the overall loss is minimized when the smallest possible number of observations is not too far to the estimate. In other words, the model will prefer making many smaller prediction errors, rather than few big ones. This feature might represent a disadvantage in the case when the true mean is closer to $\max(y) or \min(y)$ than $(\max(y) - \min(y))/2$, but still the dataset contains considerable bias with high $\Gamma$, requiring large losses in the training sample. To remedy that, I propose a new robustness set $\mathcal{S}_{(\Gamma, d)} (P, Q_X)$ and a new loss function which minimize the worst case among the new set. To do so, I introduce an additional meta-data parameter $d$, indicating direction of the bias, and indicates weather the researcher expects the true population mean to be higher or lower than the sample mean. I call this approach directional Rockafellar Uryasev regression (dRU).

\section{Directional Rockafellar Uryasev regression}

Define a robustness set $\mathcal{S}_{(\Gamma, d)}(P, Q_X)$ where \begin{align} \label{eq:gamma-biased}
    \Gamma^{-1} \leq \frac{dQ_{Y|X}(y)}{dP_{Y|X}(y)} \leq \Gamma
\end{align} and \begin{align} \label{direction}
    \mathbb{E}[Q_{Y|X}] \neq \mathbb{E}[P_{Y|X}], \quad 
    \text{sign}(\mathbb{E}[Q_{Y|X}] - \mathbb{E}[P_{Y|X}]) = \text{sign}(d).
\end{align}
Here, $d \in {-1, 1}$ is a directional parameter which indicates whether the researcher expects the target or population mean to be higher ($d = 1$) or lower ($d=-1$) than the sample mean. Due to \ref{eq:gamma-biased}, the worst case will have distribution ratio equal to either $\Gamma$ or $\Gamma^{-1}$. In this case, the worst case distribution is a distribution which assigns $\Gamma$ distribution ratio to points above the sample mean, in the case of $d = 1$, or below it in the case of $d = -1$, and $\Gamma^{-1}$ everywhere else. See Figure \ref{fig:worst_case} for a representation of one possible worst case distribution $dQ^+_{Y|X}(y)$, where the population mean is higher than the sample mean (undersampling, $d = 1$). 
To ensure that the worst case Q is a valid density function, so we can solve for $\eta$ in the following equation to find out what fraction of points can have density ratio $\Gamma$: \begin{align*}
    \Gamma (1- \eta) + (\Gamma^{-1})\eta = 1
\end{align*}
Solving this yields the value of $\eta = \Gamma/(\Gamma + 1)$. Define therefore $\eta(\Gamma) = \frac{\Gamma}{\Gamma + 1}$. Due to \ref{direction}, we assign $\Gamma$ to points above the $1 - \eta$ quantile of all losses where $\text{sign}(h(x) - y) = \text{sign}(d)$, for a given symmetric loss $L(\cdot)$, such a squared loss $L(x, y) = (x - y)^2$. The $\Gamma$ points are assigned this way in order to ensure that the worst case is the distribution with highest loss. We define $q^{L+}_{(1 - \eta)}(X; h(x))$ as the quantile function (inverse c.d.f.) of all losses such as $h(x) - y > 0$, and the converse for $q^{L-}_\eta(X; h(x))$. 
Therefore, we define the directional loss as follows, choosing the case of $d = 1$ for illustration and starting from equation 14 in \citep{sahoo2022learning}: 
\begin{align*}
    \sup\{ & \mathbb{E}_{Q_{Y|X}}[L(h(X),Y)|X=x ]: Q \in \mathcal{S}_{(\Gamma, d)}(P, Q_X) \} \\ 
    = & \mathbb{E}_{P_{Y|X}}\Big[L(h(X),Y) \Big(\Gamma^{-1} + (\Gamma + \Gamma^{-1}) \\ 
    &  \mathbb{I} \big( L(h(X), Y) \geq q^{L+}_{(1 - \eta)}(X; h(x)) \mathbb{I}(h(x) > y) \big) \Big) |X=x \Big] \\ 
    = & \Gamma^{-1}\mathbb{E}_{P_{Y|X}}\Big[L(h(X),Y)  |X=x \Big] + \\ 
    & (\Gamma + \Gamma^{-1}) \mathbb{E}_{P_{Y|X}}\Big[L(h(X),Y) \\
    & \mathbb{I} \big( L(h(X), Y) \geq q^{L+}_{(1 - \eta)}(X; h(x))\big)  \mathbb{I}(h(x) > y)  |X=x \Big]
\end{align*}
Now define \begin{align*}
    L_+(h(X), Y) & =  L(h(X), Y) \mathbb{I}(h(x) - y > 0) \\
    L_-(h(X), Y) & =  L(h(X), Y) \mathbb{I}(h(x) - y < 0) \\
    & \text{so that:}  \\
    L(h(X), Y) & = L_+(h(X), Y)  \\
    & + L_-(h(X), Y) \mathbb{I} \big(L(h(X), Y) \geq q^{L+}_{(1 - \eta)}(X; h(x)) \big) \\
    & \text{and} \\
    & \mathbb{I}(h(x) > y) = \mathbb{I} \big( L_+(h(X), Y) \geq q^{L+}_{(1 - \eta)}(X; h(x)) \big)  
\end{align*}
In this way: 
\begin{align*}
    & \Gamma^{-1}\mathbb{E}_{P_{Y|X}}\Big[L\big(h(X),Y\big)  |X=x \Big] + \\ 
    & (\Gamma + \Gamma^{-1}) \mathbb{E}_{P_{Y|X}}\Big[L\big(h(X),Y\big) \mathbb{I}(L_+(h(X), Y) \geq q^{L+}_{(1 - \eta)}(X; h(x))) |X=x \Big] \\ 
     & = \Gamma^{-1}\mathbb{E}_{P_{Y|X}}\Big[L(h(X),Y)  |X=x \Big] + \\ 
    & (\Gamma + \Gamma^{-1}) \mathbb{E}_{P_{Y|X}}\Big[ \big(L_+(h(X), Y) \\
    & + L_-(h(X), Y) \big) \mathbb{I} \big( L_+(h(X),Y) \geq q^{L+}_{(1 - \eta)}(X; h(x)) \big) |X=x \Big] \\ 
    & = \Gamma^{-1}\mathbb{E}_{P_{Y|X}}\Big[L(h(X),Y) |X=x \Big] + \\ 
    & (\Gamma + \Gamma^{-1}) \mathbb{E}_{P_{Y|X}}\Big[ L_+(h(X), Y)  \\
    & \mathbb{I} \big( L_+(h(X),Y) \geq q^{L+}_{(1 - \eta)}(X; h(x)) \big) + 0 |X=x \Big] \\
    & = \Gamma^{-1}\mathbb{E}_{P_{Y|X}}\Big[L(h(X),Y)  |X=x \Big] \\
    & + (\Gamma + \Gamma^{-1}) \eta(\Gamma) \text{CVaR}_{(1 - \eta)}(L_+(h(x), y)) 
\end{align*} where the last step is due the expectation of the indicator function, as we multiply by the expected value of $1 - P(L_+(h(X),Y) \geq q^{L+}_{(1 - \eta)}(X; h(x)))$, and plugging in the CVaR formula \begin{align*}
    \text{CVaR}_{\eta}(W) = \mathbb{E}[W | W > q_w(\eta)]
\end{align*}. 
Then, continuing: \begin{align*}
    & \Gamma^{-1}\mathbb{E}_{P_{Y|X}}\Big[L(h(X),Y)  |X=x \Big] \\
    & + (\Gamma + \Gamma^{-1}) \eta(\Gamma) \text{CVaR}_{(1 - \eta)}(L_+(h(x), y)) \\ 
    = & \Gamma^{-1}\mathbb{E}_{P_{Y|X}}\Big[L(h(X),Y)  |X=x \Big] + (\Gamma - 1) \text{CVaR}_{(1 - \eta)}(L_+(h(x), y)) \\ 
    = & \Gamma^{-1}\mathbb{E}_{P_{Y|X}}\Big[L(h(X),Y)  |X=x \Big] + \\
    & (\Gamma - 1) \bigg( \alpha(x) + \bigg( \frac{\Gamma + 1}{\Gamma} \bigg) \mathbb{E}_{P_{Y|X}}\Big[(L_+(h(X),Y) - \alpha(x))_+ |X=x \Big]   \bigg)  \\ 
    = & \Gamma^{-1}\mathbb{E}_{P_{Y|X}}\Big[L(h(X),Y)  |X=x \Big] + \\ 
    & (\Gamma - 1)\alpha(x) + \frac{\Gamma^2 - 1}{\Gamma} \mathbb{E}_{P_{Y|X}}\Big[(L_+(h(X),Y) - \alpha(x))_+ |X=x \Big].
\end{align*}
Which gives way to the directional RU loss: \begin{align*}
(h^*_{(\Gamma, d)}, a^*_{(\Gamma, d)}) & = \argmin_{h, a} \mathbb{E}_P \Big[L^\Gamma_{\text{dRU}} \Big(h(X), \alpha(X), Y \Big) \Big] \\
L^\Gamma_{\text{dRU}} (z, a, y) & = \Gamma^{-1} L(z, y) + (\Gamma - 1)a + \\ 
& \frac{\Gamma^2 -1}{\Gamma}(L(z, y) > a)_+\mathbb{I}(sign(h(x) - y) = sign(d))
\end{align*}

\begin{figure}[h]
    \centering
    \includegraphics[trim=0 0 0 80, clip, width=1\linewidth]{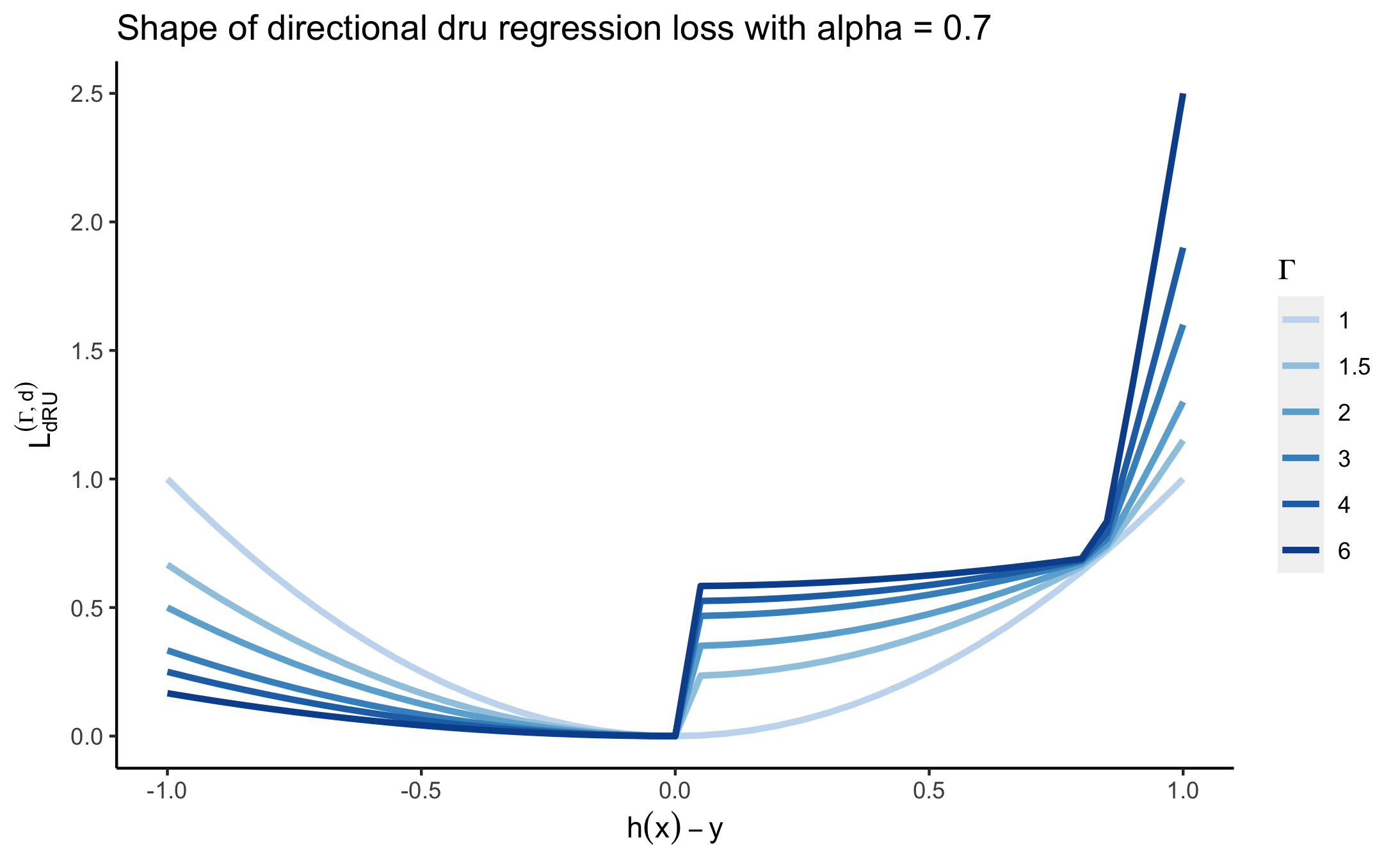}
    \caption{Visualization of approximate dRU regression loss for different values of $\Gamma$, $d = 1$}
    \label{fig:dloss_viz_gamma}
\end{figure}

\begin{figure}[h]
    \centering
    \includegraphics[trim=0 0 0 80, clip, width=1\linewidth]{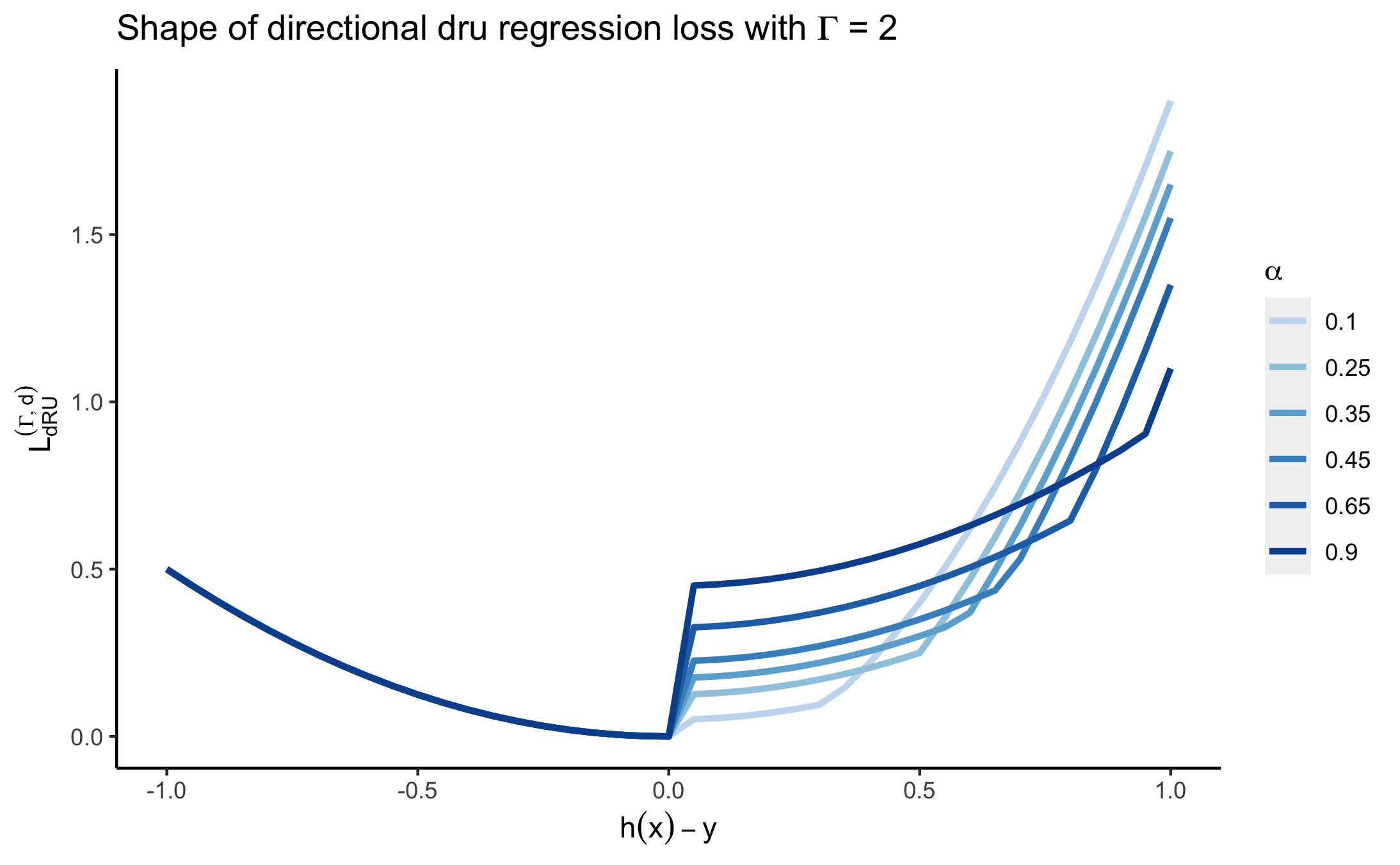}
    \caption{Visualization of approximate dRU regression loss for different values of $a$, $d = 1$}
    \label{fig:dloss_viz_alpha}
\end{figure}

A schematic visualization of the approximate loss is provided in Figures \ref{fig:dloss_viz_gamma} and \ref{fig:dloss_viz_alpha}. 

With the additional parameter, the penalization for larger losses is applied only in the case of $\text{sign}(h(x) - y) = \text{sign}(d)$. In this way, the model can estimate closer to $\max(y) or \min(y)$ even in the case of high $\Gamma$. The additional $d$ parameter indicates whether the data is expected to be under or over sampled. In other words, $d$ encourages the model to avoid over-estimations or under-estimations, depending on the sign. 

\section{Justification}

Researchers often might have an understanding of weather their data contain bias (Gamma) and of the direction of the bias (over or under estimation). For example, election polls constantly underestimate the right-wing conservative parties, as well as over-estimate populist parties \citep{selcuki2023turkishpolls, dickie2020polls}. Field experts that rely on the same panel for estimation election after election might be aware of these inherit bias in the data, but might have no direct way to add this information to estimation. Bayesian approaches allow researchers to add prior knowledge on the distribution of coefficients or intercept of a regression model. Nonetheless, when $P(Y|X, S) \neq P(Y|X)$ it is hard to identify a single coefficient to correct, or just change the intercept. Therefore, researchers could benefit for a more straightforward way to add meta-information in the analysis of data.  

\section{Application} 
\begin{figure}[h]
    \centering
    \includegraphics[width=\linewidth]{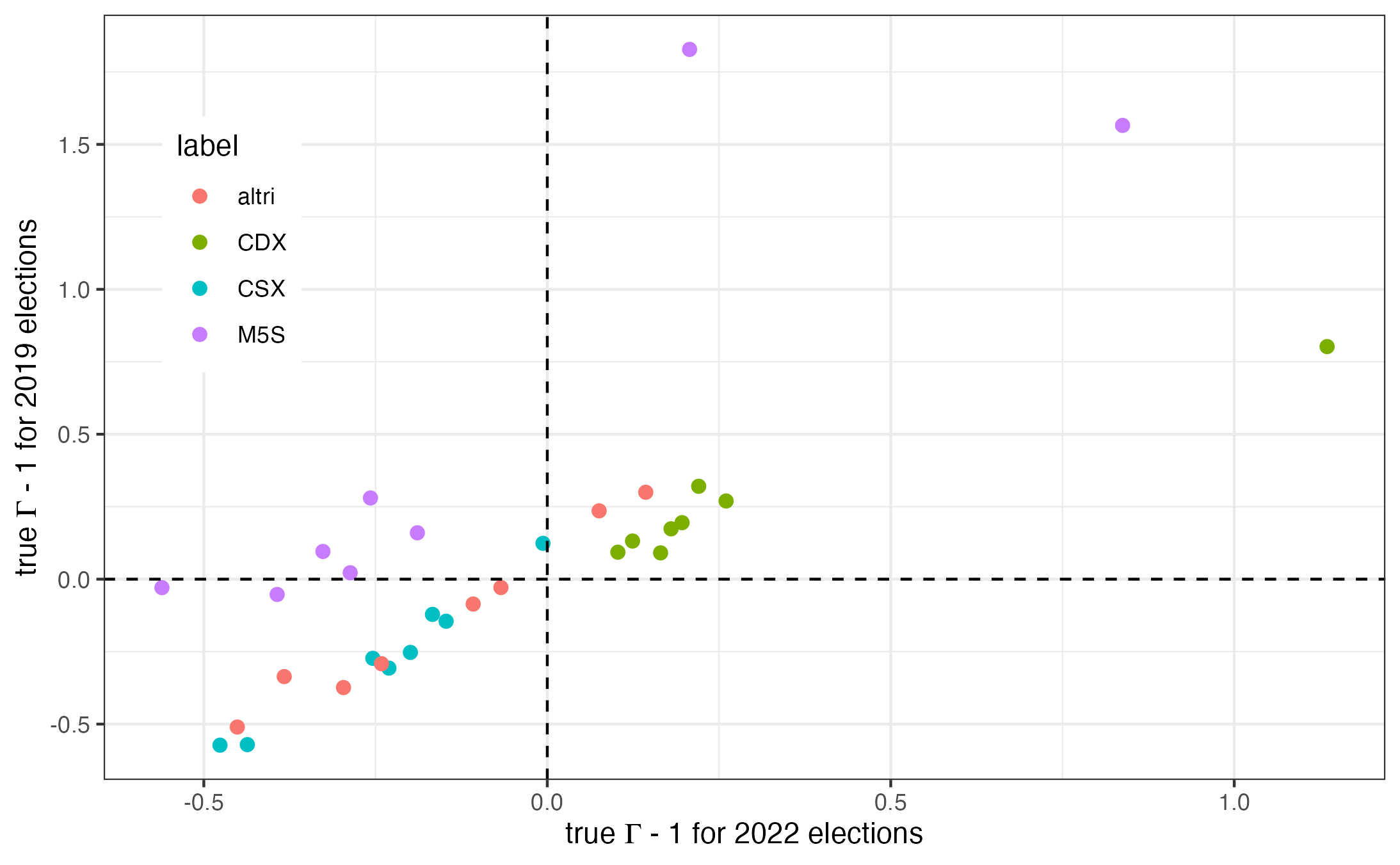}
    \caption{In-sample correlation between true $\Gamma$ and $d$ values in two consecutive Italian national elections.}
    \label{fig:2022vs2018}
\end{figure}

I test the model in an applied setting constructed so to resemble a situations researchers might find themselves into. In the scenario, 5 non-probability datasets are used to predict the results of the 2022 Italian national elections. Datasets were collected by Demetra Opinioni s.r.l., as a mixture of random digit dialing and use of their proprietary online panel. The online panel was known to already contain selection bias from previous results \citep{mixmode19demetra}. Details on the data and training procedure can be found in Section \ref{Appendix}. The covariates available are gender, age, geographical area, employment and education status and vote on previous election. A standard practice in electoral estimation is to use the national census to re-weight or model the data in accordance to the general population, a method called post-stratification \citep{si2020use}. Nonetheless, the Italian census is not available for all the cross-tabulated combinations of variables, but only for selected combinations. Moreover, the lack of publicly available exit polls means that only the marginal distribution of the past vote is available. It is clear how in this setting the available covariates do not suffice to fully explain sample inclusion probabilities. We aim to test different estimation strategies for this scenario. In this case, training will consist in selecting one of the datasets together with a set of covariates for estimation. The model is trained and then tested against the ground truth election results. The process is repeated for all datasets and all sets of covariates, permuting all possible options, corresponding to all possible variable selection decisions. For each permutation a score $b \in (-\infty, 1]$ indicates the amount of bias added or removed by the estimation method. The score is calculated as follows: \begin{align*}
    b = \sum_{i}^{i \in \text{party}} \frac{|\bar{y}^{\text{true}}_i - \bar{y}^{\text{unweighted}}_i| - |\bar{y}^{\text{true}}_i - \hat{y}_i|}{|\bar{y}^{\text{true}}_i - \bar{y}^{\text{unweighted}}_i|}
\end{align*} 
Each neural network is tasked with estimating the population average of each of the 5 main political coalitions in the 2022 Italian elections. Since more than 14 individual parties or conglomerates were participating in the elections, they were aggregated into 5 politically coherent coalitions in order to ease computational cost. See Section \ref{Appendix} for more details on the parties aggregation. For each estimation method the distribution of $b$-scores is compared. Positive $b$ values indicate the fraction of bias that the method eliminated, across all political parties, compared to using a simple average to predict population mean values. A value of $b = 1$ indicates that the estimation method predicted the election outcome with perfect precision. A value of $b = 0$ indicates that the estimation method was not to remove any bias overall. A negative value of $b$ indicates that the estimation method made prediction worse off. This setting represents the expected outcome of applying the corresponding method to a non-probability dataset with a set of variables. To inform the choice of hyper-parameter vectors $\Gamma$ and $d$ we use the data from the previous elections, the 2018 National Election, indicating the true values of $\Gamma$ and $d$ of the $Q$ distributions in the previous case for the sample panel for each party. We find the true $\Gamma$ and $d$ values for the 2018 election to reliably correlated with the true values for the 2022 election in the majority cases. The true values are plotted in Figure \ref{fig:2022vs2018}. In our application, we compare the following estimation methods: 
\begin{enumerate}
    \item dRU regression with $\Gamma$ and $d$ values informed from the in-sample past vote distribution; 
    \item NN: dRU with $d = 0$ and $\Gamma = 1$ representing the case of data expected to me MAR (no added meta-information);
    \item MRP: Multilevel regression and post-stratification, one of the most common approaches in election studies \citep{gelman1997poststratification} which nonetheless assumes MAR;
    \item pinball, a neural network with a pinball loss in the form of \begin{align*}
    f \big( h(x), y, p \big) =
    \begin{cases}
    p \big( h(x) - y \big)^2 & \text{if } h(x) > y \\
    (1-p) \big( h(x) - y \big)^2 & \text{if } h(x) < y
    \end{cases}
    \end{align*} A pinball loss estimates the $p$-th quantile and can therefore can also be used to nudge the estimate upward or downward depending on the expectations. This represents a similar functioning to dRU, and therefore can be considered a close competitor. The $p$-hyperparameter values are chosen based on previous election results true bias, similarly to dRU.
    \item dRU with swapped $\Gamma$, where the vector of $\Gamma$ values is inverted in order. This displays the case where meta-information on the quantity of bias is wrong. 
    \item dRU with swapped sign of $d$, to display the case where meta-information on direction of bias is wrong. 
    \item dRU with both $\Gamma$ and $d$ swapped, indicating the case where the meta-information provided to the model is most probably wrong. 
\end{enumerate}
 
\begin{figure}[h]
    \centering
    \includegraphics[width=\linewidth]{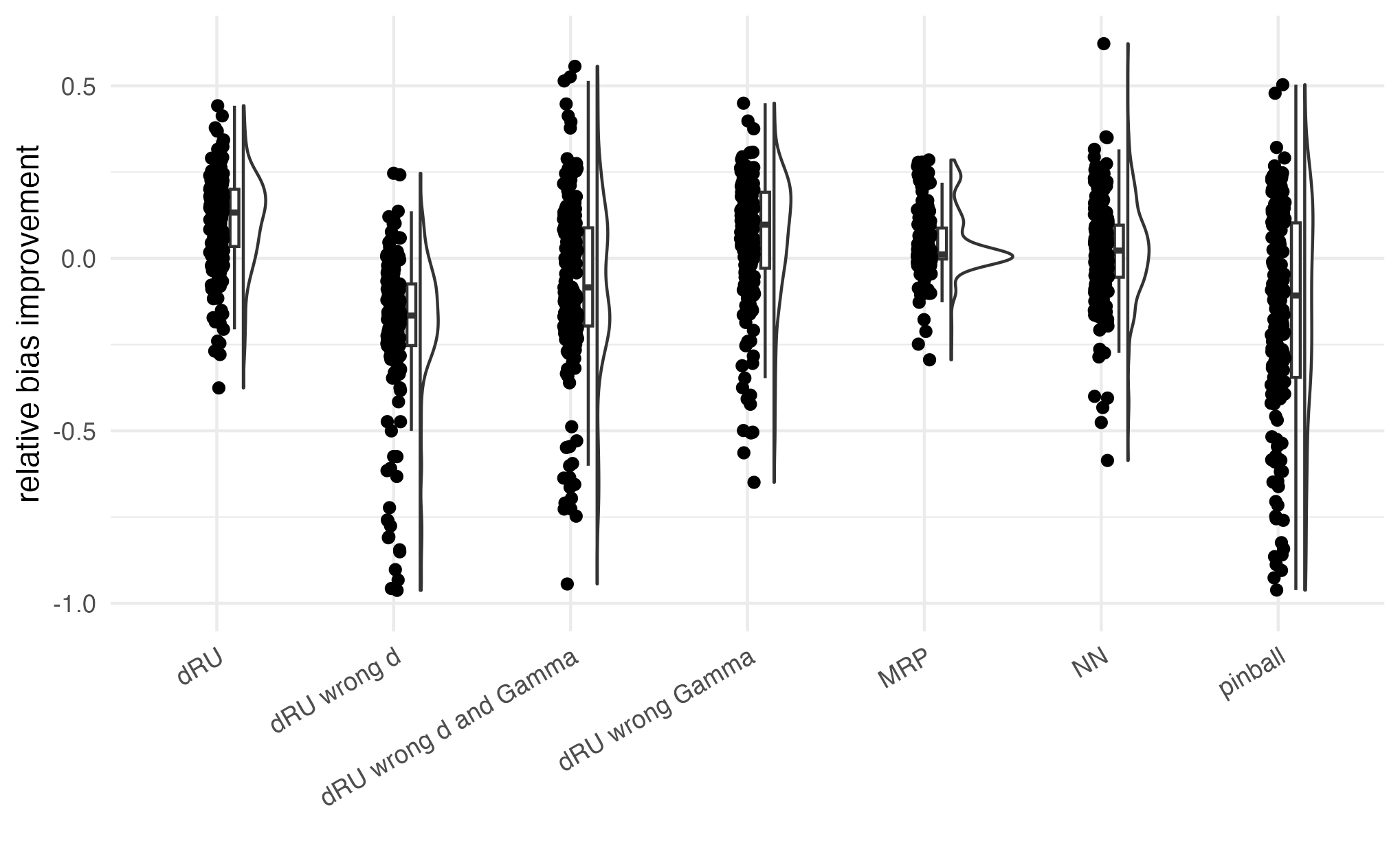}
    \caption{Distribution of $b$-scores for all estimation methods}
    \label{fig:results}
\end{figure}

 The results are displayed in Figure \ref{fig:results} and in table \ref{tab:table_results}. 

\begin{table}[htbp]
  \centering
    \begin{tabular}{|l|l|l|}
    \hline
    \textbf{Estimation Method} & \textbf{Average b-score} & \textbf{freq. b $>$ 0} \\
    \hline
    MRP              & 0.0419            & 0.6942            \\
    NN               & 0.0180            & 0.5777          
    \\
    dRU              & 0.1090            & 0.8204            \\
    dRU wrong Gamma  & 0.0571            & 0.6893            \\
    dRU wrong d      & -0.2181           & 0.1165            \\
    dRU wrong d and Gamma & -0.0956     & 0.4126            \\
    \hline
    \end{tabular}%
    \caption{Average $b$-score and average frequency of positive bias removal ($b > 0$) for each estimation method }
  \label{tab:table_results}%
\end{table}%

\section{Discussion}
The results indicate that dRU regression, when equipped with $d$ and $\Gamma$ parameter extracted from the in-sample distribution of previous election results, provides the best overall reduction of bias compared to similar methods. Changing the direction of bias or the amount of $\Gamma$ worsen the performance of the model and adds variability to the estimate. It can be seen how a wrong $d$ can have a worse effect than a wrong $\Gamma$. NN shows the performance of the model when $\Gamma = 1, d = 0$. In that case the distribution of $b$-scores is more variable, which is to be expected since the datasets are MNAR. Similarly, MRP has an overall mixed performance, being able to reduce the bias for the majority of cases, but worsening the situation in others. Finally, the pinball loss neural network's performance results inferior to the others in therms of variability and bias reduction. 
 
\section{Conclusion}
The present paper proposes a new loss function, dRU, based on recent advances in DRO. The new loss function allows researchers to modulate estimation in order to account both for quantity of bias, expressed by $\Gamma \leq 1$ and direction of bias (over or under estimation) expressed by $d \in \{-1, 1\}$. The loss function allow to obtain a decision rule $h(x)$ which minimizes the worst case among a robustness set where the sample has been generated by a population with a sampling process of bias up to $\Gamma$, where the population mean is higher (or lower) compared to the sample mean, depending on the sign of $d$. The present paper compares the estimation with dRU against close competitors, showing how dRU produces the best and most frequent reduction in bias. Therefore, it can be concluded that the possibility to include this meta-information increases the quality of estimates drawn from a non-probability election dataset. Some researcher have humorously stated that non-probability, biased sample are "almost everywhere" and there is no such thing as a probability, representative sample. While the spread of non-random samples such as Big Data might cause an crisis in estimation reliability, the question remain on what to do with this data once is collected. While drawing estimates with biased data will mostly produced erroneous model, the estimation can be adjusted considering meta-information available to the researcher. Such approach will allow researcher to better leverage on the often rich quantity of expertise on the sampling process or on the subject at hand. I believe such approach can be especially useful in case of repeated measurements through time, especially in the case of online longitudinal panels or electoral polling. 

\section*{Acknowledgments}
I would like to thank Ms. Roshni Sahoo for the fruitful explanations of the Rockafellar-Uryasev regression derivation. I would like to thank Prof. Yajuan Si, Prof. Maria Letizia Tanturri and Prof. Omar Paccagnella for their continuous support in the development of the ideas in this paper. 



\section{Appendix: data collection and model training methods} \label{Appendix}

\subsection{Data collection}
Data consisted of five non-probability datasets, total $N = 16747$. All data was originally conducted by Demetra Opinioni.net. The company operates a proprietary online panel where respondents can receive small monetary incentives to complete polls. For the current datasets, in waves 1 to 3 individuals were contacted across the panel using a quota sampling method. For waves 1 and 2, in addition to this form of CAWI sampling, additional individuals were contacted through CATI, in a random number dialling. Wave 3 and postwave are online panel only. For postwave, it was collected shortly after the election results, and was proposed to the entirety of the web panel, in order to draw more responses. Therefore, no quotas were allocated in the sampling. Finally, wavesm, which stands for social media, was collected some time later, with the purpose of also having a social media sample together with the online panels one. The sample was collected through advertisements on Meta's platforms, Facebook and Instagram. The respondents did not get any monetary incentive, and were recruited through Meta's algorithm for ad placement. A pseudo-quota mechanism was implemented so that different ads were created for different age targets. This was done in order to collect some respondents in the younger cells of the population, which are generally considered harder to recruit on these social media platforms \citep{kuhne2020using}. 

\begin{table}[htbp]
\centering
\begin{tabular}{cccccc}
\toprule
& \textbf{Wave1} & \textbf{Wave2} & \textbf{Wave3} & \textbf{Wavepost} & \textbf{Wavesm} \\
\midrule
\textbf{N} & 911 & 966 & 1736 & 10147 & 1289 \\
\textbf{Mode} & Mixed & Mixed & Cawi & Cawi & Social Media \\
\textbf{Date} & 08-2022 & 09-2022 & 09-2022 & 10-2022 & 07-2023 \\
\bottomrule
\end{tabular}
\caption{Description of Waves}
\label{tab:waves}
\end{table}

\subsection{Questionnaires}
While the content of the questionnaires is not entirely reported here for reasons of brevity, a set of common questions were asked across all questionnaires. For a schematic description of the included categorical variables and their corresponding number of levels, see table \ref{tab:vars}. These variables were selected as the only ones for which census cross-population totals were available. This is an usual requirement for post-stratification \citep{si2020use}. All variables were present in all datasets, with the exception of wavepost, the largest survey, for which the 2018 vote preference was not recorded. 

\begin{table}[htbp]
\centering
\begin{tabular}{ccccccc}
\toprule
& \textbf{Gender} & \textbf{Age} & \textbf{Area} & \textbf{Education} & \textbf{Employment} & \textbf{2018 Vote}\\
\midrule
 & 2 & 5 & 4 & 3 & 3 & 4\\
\bottomrule
\end{tabular}
\caption{Variables with corresponding number of levels }
\label{tab:vars}
\end{table}

\subsection{Political aspects during the 2022 elections}
The 2022 elections resulted in a victory for the right-wing coalition (CDX), which obtained the majority across most provinces, and in the surprising collapse for the Movimento 5 Stelle party (M5S). The centre-left coalition (CSX) also suffered with poor results across the board. Notably, the 2022 elections saw a relatively large share of votes towards an emergent party, for which no previous information could be obtained from past elections: a center progressist party named Azione-Italia Viva (Az), which has also been called Third-Pole. The party did not obtain the majority in any province, but obtained sensible results in some metropolitan centers.

\subsection{Data processing}
The data was imported in SPSS format into RStudio. Then, for each dataset, all rows where at least one of the common categorical variables or the response variables were missing (Table \ref{tab:vars} for a list) were dropped. Table \ref{tab:waves} contains description values of the processed dataset after removing for this non-response. Then, voting preferences of the respondents were aggregated into political coalitions for each of the 2028 and 2022 expressed preferences. This aggregation was performed in order to reduce the number of parties and simplify further applications. The aggregation resulted into five larger coalitions: CDX (right-wing coalition), CSX (left-wing coalition), M5S (movimento 5 stelle), Az (Azione - Italia Viva) and others (all other parties). The same coalitions were also organized for the 2018 preferences, with the exception of Az which was not present at the time. Non voters included those who reported not to vote or reported to leave the voting ballot empty or white. The non voters rows were also dropped and all estimation was carried out targeting the share of votes for each coalition. 

\begin{table}[ht]
    \centering
    \caption{Political parties aggregation for the 2022 elections}
    \begin{tabular}{|l|p{0.7\linewidth}|}
        \hline
        \textbf{Coalition} & \textbf{Individual Parties} \\
        \hline
        CDX & Lega - Salvini premier, Forza Italia (Berlusconi), Fratelli d'Italia (Meloni), Noi Moderati (Noi con l'Italia, Coraggio Italia, Italia al Centro - Lupi, Toti, Brugnaro) \\
        \hline
        CSX & Partito Democratico (Letta) con Articolo Uno e PSI, +Europa (Bonino), Alleanza Verdi e Sinistra (Fratoianni e Bonelli), Civica Popolare - Lorenzin, Liberi e Uguali, Impegno Civico (Di Maio) - Centro Democratico (Tabacci) \\
        \hline
        M5S & Movimento 5 Stelle (Conte) \\
        \hline
        Az & Azione - Italia Viva (Calenda e Renzi) \\
        \hline
        altri & Italia Sovrana e Popolare (Rizzo e Ingroia), Italexit per l'Italia (Gianluigi Paragone), Unione Popolare (De Magistris), Partito Comunista - Rizzo, other party \\
        \hline
        non voters & empty vote, I would not like to vote \\
        \hline
    \end{tabular}
    \label{fig:conv_2022}
\end{table}

\begin{table}[ht]
    \centering
    \caption{Political parties aggregation for the 2018 elections}
    \begin{tabular}{|l|p{0.7\linewidth}|}
        \hline
        \textbf{Coalition} & \textbf{Individual Parties} \\
        \hline
        CDX & Lega - Salvini, Forza Italia, Fratelli d'Italia - Meloni, Noi con l'Italia - UDC \\
        \hline
        CSX & Partito Democratico-PD, Più Europa - Bonino, Insieme (Verdi, PSI, Area Civica), Civica Popolare - Lorenzin \\
        \hline
        M5S & Movimento 5 Stelle \\
        \hline
        altri & Liberi e Uguali, Potere al Popolo, Casapound Italia, Il Popolo della Famiglia - Adinolfi, Italia agli Italiani - Forza Nuova, Partito Comunista - Rizzo,  other party \\
        \hline
        non voters & empty vote, I would not like to vote \\
        \hline
    \end{tabular}
    \label{fig:conv_2018}
\end{table}

\subsection{Neural Network}
The neural network used for dRU estimation was developed using the \texttt{torch} library for R. The $h(x)$ and the $\alpha(x)$ networks where identically built with one input layer, one hidden layer and one output layer. Each layer consisted of 4 neurons, and was associated with a relu activation function. Early stopping was implemented for training, so that the each dataset was split into a 0.9 training and 0.1 validation trances. The neural net would train for 20 epochs or up until the validation loss remained flat for 3 consecutive iterations. Batch size was fixed to 12 and the learning rate to 0.01. Ad Adam optimizer was implemented for the gradient descent. The pinball loss neural network consistent of a single $h(x)$ network with the same structure but double the amount of neurons per layer. The remaining hyper-parameters where kept the same. Once training is completed, the estimates $\tilde{y}$ are subjected to post-stratification in the form of \begin{align*}
    \hat{y}_{i} = \sum_j \tilde{y}_{i,j} P(X = j)
\end{align*} where $j$ is the post-stratification cell, and $P(X = j)$ represents the fraction of the target population to belong to that population cell. For more information see \cite{gelman1997poststratification} or \cite{si2020use}. 

\bibliography{reference}

\end{document}